\documentclass[runningheads]{llncs}
\usepackage{graphicx}
\usepackage{comment}
\usepackage{amsmath,amssymb} 
\usepackage{color}
\usepackage{bm}
\usepackage{url}
\usepackage{makecell}
\usepackage{multirow}
\usepackage{varwidth}
\usepackage{cite}
\usepackage{algorithm}
\usepackage{algorithmic}
\usepackage[misc]{ifsym}
\usepackage[colorlinks,linkcolor=red,anchorcolor=blue,citecolor=green]{hyperref}

\PassOptionsToPackage{numbers, compress}{natbib}

\newcommand{\PreserveBackslash}[1]{\let\temp=\\#1\let\\=\temp}
\newcolumntype{C}[1]{>{\PreserveBackslash\centering}p{#1}}
\newcolumntype{R}[1]{>{\PreserveBackslash\raggedleft}p{#1}}
\newcolumntype{L}[1]{>{\PreserveBackslash\raggedright}p{#1}}

\newenvironment{shrinkeq}[2]
{ \bgroup
  \addtolength\abovedisplayshortskip{#1}
  \addtolength\abovedisplayskip{#1}
  \addtolength\belowdisplayshortskip{#2}
  \addtolength\belowdisplayskip{#2}}
{\egroup\ignorespacesafterend}

\begin{document}
\pagestyle{headings}
\mainmatter
\def\ECCVSubNumber{984}  

\title{\mbox{Dual Adversarial Network: Toward Real-world} Noise Removal and Noise Generation} 


\titlerunning{DANet for noise removal and generation}
%
\author{Zongsheng Yue\inst{1,2} \and
Qian Zhao\inst{1} \and
Lei Zhang\inst{2,3} \and
Deyu Meng\inst{1,4,\textrm{\Letter}}}
\authorrunning{Zongsheng Yue et al.}
%
\institute{Xi'an Jiaotong University, Shaanxi, China \\
\email{\{zsyzam,timmy.zhaoqian\}@gmail.com, dymeng@mail.xjtu.edu.cn}\and
Hong Kong Polytechnic University, Hong Kong, China\\
\email{cslzhang@comp.polyu.edu.hk} \and
DAMO Academy, Alibaba Group, Shenzhen, China \and
The Macau University of Science and Technology, Macau, China
}
\maketitle

\begin{abstract}
Real-world image noise removal is a long-standing yet very challenging task in computer vision.
The success of deep neural network in denoising stimulates the research of
noise generation, aiming at synthesizing more clean-noisy image pairs to facilitate the training of deep denoisers.
In this work, we propose a novel unified framework to simultaneously deal with the noise removal and
noise generation tasks. Instead of only inferring the posteriori distribution of the latent clean image conditioned on
the observed noisy image in traditional MAP framework, our proposed method learns the joint distribution of
the clean-noisy image pairs.
Specifically, we approximate the joint distribution with two different factorized forms, 
which can be formulated as a denoiser mapping the noisy image to the clean one and a generator mapping
the clean image to the noisy one.
The learned joint distribution implicitly contains all the information between
the noisy and clean images, avoiding the necessity of manually designing the image priors and noise assumptions
as traditional.
Besides, the performance of our denoiser can be further improved by augmenting the original training
dataset with the learned generator.
Moreover, we propose two metrics to assess the quality of the generated noisy image, for which, to the best of
our knowledge, such metrics are firstly proposed along this research line.
Extensive experiments have been conducted to demonstrate the superiority
of our method over the state-of-the-arts both in the real noise removal and generation tasks.
The training and testing code is available at \url{https://github.com/zsyOAOA/DANet}.
\end{abstract}

\section{Introduction}

Image denoising is an important research problem in low-level vision,
aiming at recovering the latent clean image $\bm{x}$ from its noisy observation $\bm{y}$.
Despite the significant advances in the past decades~\cite{buades2005non,dabov2007image,zhang2017beyond,yue2019variational},
real image denoising still remains a challenging task, due to the complicated processing steps within
the camera system, such as demosaicing, Gamma correction and compression~\cite{tsin2001statistical}.

From the Bayesian perspective, most of the traditional denoising methods can be interpreted within the Maximum A
Posteriori (MAP) framework, i.e., $\max_{\bm{x}} p(\bm{x}|\bm{y}) \propto p(\bm{y}|\bm{x})p(\bm{x})$, which involves one
likelihood term $p(\bm{y}|\bm{x})$ and one prior term $p(\bm{x})$. Under this framework,
there are two methodologies that have been considered.
The first attempts to model the likelihood term with proper distributions,
e.g., Gaussian, Laplacian, MoG~\cite{Meng_2013_ICCV,zhu2016blind,yue2019robust} and MoEP~\cite{Cao_2015_ICCV},
which represents different understandings for the noise generation mechanism, while the second mainly focuses on
exploiting better image priors, such as total variation~\cite{rudin1992nonlinear}, non-local
similarity~\cite{buades2005non}, low-rankness~\cite{dong2012nonlocal,gu2014weighted,wang2018weakly,MCWNNM} and
sparsity~\cite{mairal2007sparse,zhou2009non,Xu_2018_ECCV}. Despite better interpretability led
by Bayesian framework, these MAP-based methods are still limited by the manual assumptions on the noise and
image priors, which may largely deviate from the real images.
\begin{figure}[t]
    \centering
    \includegraphics[scale=0.57]{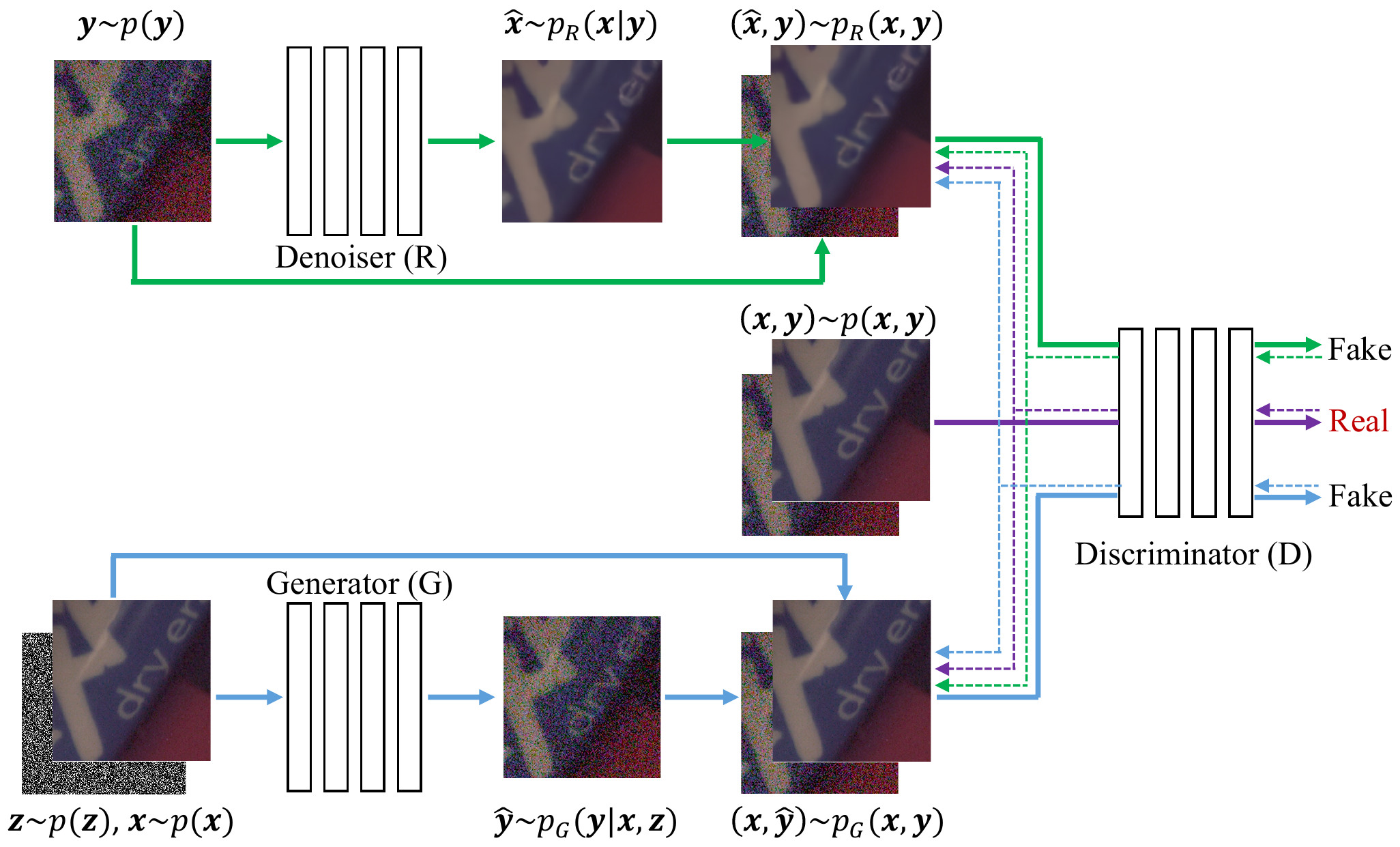}
    \vspace{-2mm}
    \caption{Illustration of our proposed dual adversarial framework. The solid lines denote the forward
        process, and the dotted lines mark the gradient interaction between the denoiser and generator during the
        backword. }
    \label{fig:model-framework}
\end{figure}

In recent years, deep learning (DL)-based methods have achieved impressive success in image denoising
task~\cite{zhang2017beyond,anwar2019real,yue2019variational}. However, as is well known,
training a deep denoiser requires large amount of clean-noisy image pairs, which are time-consuming and expensive
to collect. To address this issue, several noise generation\footnote{ The phrase ``noise generation" indicates
the generation process of noisy image from clean image throughout this paper.}
approaches were proposed to simulate more clean-noisy image pairs to facilitate the training of deep denoisers. The main
idea behind them is to unfold the in-camera processing pipelines~\cite{guo2019toward,brooks2019unprocessing},
or directly learn the distribution $p(\bm{y})$ as in ~\cite{chen2018image,kim2019grdn} using generative adversarial
network (GAN)~\cite{goodfellow2014generative}.
However, the former methods involve many hyper-parameters needed to be carefully tuned for specific cameras,
and the latter ones suffer from simulating very realistic noisy image
with high-dimensional signal-dependent noise distributions.
Besides, to the best of our knowledge, there is still no metric to quantitatively assess the quality
of the generated noisy images w.r.t. the real ones.

Against these issues, we propose a new framework to model the joint distribution $p(\bm{x},\bm{y})$ instead
of only inferring the conditional posteriori $p(\bm{x}|\bm{y})$ as in conventional MAP framework. Specifically,
we firstly factorize the joint distribution $p(\bm{x},\bm{y})$ from two opposite directions,
i.e., $p(\bm{x}|\bm{y})p(\bm{y})$ and $\int_{\bm{z}}p(\bm{y}|\bm{x},\bm{z})p(\bm{x})p(\bm{z})\mathrm{d}\bm{z}$,
which  can be well approximated by a image denoiser and a noise generator. Then we simultaneously train the denoiser 
and generator in a dual adversarial manner as illustrated in Fig.~\ref{fig:model-framework}. After that, the learned denoiser can either be directly
used for the real noise removal task, or further enhanced with new clean-noisy image pairs simulated by the learned generator.
In summary, the contributions of this work can be mainly summarized as:
\begin{itemize}
    \item Different from the traditional MAP framework, our method approximates
        the joint distribution $p(\bm{x},\bm{y})$ from two different factorized forms in a dual adversarial manner,
        which subtlely avoids the manual design on image priors and noise distribution. What's more, the joint distribution
        theoretically contains more complete information underlying the data set comparing with the conditional posteriori $p(\bm{x}|\bm{y})$.
    \item Our proposed method can simultaneously deal with both the noise removal and noise generation tasks in
        one unified Bayesian framework, and achieves superior performance than the state-of-the-arts
        in both these two tasks. What's more, the performance of our denoiser can be further improved
        after retraining on the augmented training data set with additional clean-noisy image pairs simulated
        by our learned generator.
    \item In order to assess the quality of the simulated noisy images by a noise generation method, we design
        two metrics, which, to the best of our knowledge, are the first metrics to this aim.
\end{itemize}

\section{Related Work}
\subsection{Noise Removal}
Image denoising is an active research topic in computer vision. Under the MAP framework, rational priors
are necessary to be pre-assumed to enforce some desired properties of the recovered image.
Total variation~\cite{rudin1992nonlinear}
was firstly introduced to deal with the denoising task. Later, the non-local similarity prior, meaning that
the small patches in a large non-local area may share some similar patterns,
was considered in NLM~\cite{buades2005non} and followed by many other denoising
methods~\cite{dabov2007image,dong2012nonlocal,Mairal2009,Lebrun2013}.
Low-rankness~\cite{dong2012nonlocal,gu2014weighted,yong2017robust,MCWNNM} and
sparsity~\cite{mairal2007sparse,xie2017kronecker,Mairal2009,zhou2009non,Xu_2018_ECCV} are another two well-known image priors, which are
often used together within the dictionary learning methods. Besides, discriminative learning methods also represent
another research line, mainly including Markov random field (MRF) methods~\cite{Barbu2009,Samuel2009,Sun2011},
cascade of shrinkage fields (CSF) methods~\cite{Schmidt2014,Schmidt2017} and the trainable nonlinear reaction
diffusion (TNRD)~\cite{Chen2017} method. Different from above priors-based methods, noise modeling approaches
focus on the other important component of MAP, i.e., likelihood or fidelity term. E.g., Meng and
De La Torre~\cite{Meng_2013_ICCV}
proposed to model the noise distribution as mixture of Gaussians (MoG), while
Zhu \textit{et al.}~\cite{zhu2016blind} and Yue \textit{et al.}~\cite{yue2019robust} both introduced the
non-parametric Dirichlet Process to MoG to expand its flexibility. Furthermore,
Cao \textit{et al.}~\cite{Cao_2015_ICCV} proposed the mixture of expotential
power (MoEP) distributions to fit more complex noise.

In recent years, DL-based methods achieved significant advances in the image denoising task. Jain and
Seung~\cite{Jain2008} firstly adopted a five-layer network to deal with the denoising task. Then Burger
\textit{et al.}~\cite{Burger2012} obtained the comparable performance with BM3D using one plain multi-layer
perceptron (MLP). Later, some auto-encoder based methods~\cite{Xie2012,Agostinelli2013} were also immediately
proposed. It is worthy mentioning that Zhang \textit{et al.}~\cite{zhang2017beyond} proposed the convolutional
denoising network DnCNN and achieved the state-of-the-art performance on Gaussian denoising. Following DnCNN,
many different network architectures were designed to deal with the denoising task, including 
RED~\cite{Mao2016}, MemNet\cite{Tai2017}, NLRN~\cite{Liu2018}, N3Net~\cite{Ploetz2018},
RIDNet~\cite{anwar2019real} and VDN~\cite{yue2019variational}.

\subsection{Noise Generation}
As is well known, the expensive cost of collecting pairs of training data is a critical limitation
for deep learning based denoising methods. Therefore, several methods were proposed to explore the generation
mechanism of image noise to facilitate an easy simulation of more training data pairs. 
One common idea was to generate image pairs
by ``unprocessing" and ``processing" each step of the in-camera processing pipelines,
e.g.,~\cite{guo2019toward,brooks2019unprocessing,Jaroensri2019}. However, these methods involve many
hyper-parameters to be tuned for specifi camera. Another simpler way was to learn the real
noise distribution directly using GAN~\cite{goodfellow2014generative} as demonstrated in~\cite{chen2018image}
and~\cite{kim2019grdn}. Due to the complexity of real noise and the instability of training GAN, it is very
difficult to train a good generator for simulating realistic noise.

\section{Proposed Method}
Like most of the supervised deep learning denoising methods, our approach is built on the given training data set
containing pairs of real noisy image $\bm{y}$ and clean image $\bm{x}$, which are accessible thanking to the contributions
of~\cite{Anaya2014,SIDD_2018_CVPR,xu2018real}. Instead of forcely learning a mapping
from $\bm{y}$ to $\bm{x}$, we attempt to approximate the underlying joint distribution $p(\bm{x},\bm{y})$ of
the clean-noisy image pairs. In the following, we present our method from the Bayesian perspective.

\subsection{Two Factorizations of Joint Distribution}\label{subsec:joint-sampling}
In this part, we factorize the joint distribution $p(\bm{x},\bm{y})$ from two different perspectives, and
discuss their insights respectively related to the noise removal and noise generation tasks.

\vspace{2mm}
\noindent\textbf{Noise removal perspective:} The noise removal task can be considered as inferring the conditional
distribution $p(\bm{x}|\bm{y})$ under the Bayesian framework. The learned denoiser $R$ in this task represents
an implicit distribution $p_R(\bm{x}|\bm{y})$ to approximate the true distribution $p(\bm{x}|\bm{y})$. 
The output of $R$ can be seen as an image sampled from this implicit distribution $p_R(\bm{x}|\bm{y})$.
Based on such understanding, we can obtain a pseudo clean image pair $(\hat{\bm{x}}, \bm{y})$ as
follows\footnote{We mildly assume that $\bm{y} \sim p(\bm{y})$
is easily implemented by sampling $\bm{y}$ from the empirical distribution $p(\bm{y})$ of the training
data set, and so does as $\bm{x} \sim p(\bm{x})$.}, i.e.,
\begin{equation}
     \bm{y} \sim p(\bm{y}),~\hat{\bm{x}} = R(\bm{y})\Longrightarrow (\hat{\bm{x}}, \bm{y}),
    \label{eq:sampling-D}
\end{equation}
which can be seen as one example sampled from the following pseudo joint distribution:
\begin{equation}
    p_R(\bm{x},\bm{y}) = p_R(\bm{x}|\bm{y})p(\bm{y}).
    \label{eq:joint-factor-D}
\end{equation}
Obviously, the better denoiser $R$ is, the more accurately that the pseudo joint distribution $p_R(\bm{x},\bm{y})$
can approximate the true joint distribution $p(\bm{x},\bm{y})$.

\vspace{2mm}
\noindent \textbf{Noise generation perspective:}
In real camera system, image noise is derived from multiple hardware-related random noises (e.g., short noise, thermal noise), and further affected by in-camera processing
pipelines (e.g., demosaicing, compression). After introducing an additional latent variable $\bm{z}$, representing the fundamental elements conducting the hardware-related
random noises, the generation process from $\bm{x}$ to $\bm{y}$ can be depicted by the conditional distribution
$p(\bm{y}|\bm{x},\bm{z})$.
The generator $G$ in this task expresses an implicit distribution $p_G(\bm{y}|\bm{x},\bm{z})$
to approximate the true distribution $p(\bm{y}|\bm{x},\bm{z})$. The output of $G$ can be seen as an example sampled
from $p_G(\bm{y}|\bm{x},\bm{z})$, i.e., $G(\bm{x},\bm{z})\sim p_G(\bm{y}|\bm{x},\bm{z})$.
Similar as Eq.~\eqref{eq:sampling-D}, a pseudo noisy image pair $(\bm{x},\hat{\bm{y}})$ is easily obtained:
\begin{equation}
    \bm{z} \sim p(\bm{z}),~ \bm{x} \sim p(\bm{x}), ~ \hat{\bm{y}} = G(\bm{x}, \bm{z}) 
    \Longrightarrow (\bm{x}, \hat{\bm{y}}),
    \label{eq:sampling-G}
\end{equation}
where $p(\bm{z})$ denotes the distribution of the latent variable $\bm{z}$, which can be easily set as an isotropic
Gaussian distribution $\mathcal{N}(0, \bm{I})$.

Theoretically, we can marginalize the latent variable $\bm{z}$ to obtain 
the following pseudo joint distribution $p_G(\bm{x},\bm{y})$ as an approximation to 
$p(\bm{x},\bm{y})$:
\begin{equation}
    p_G(\bm{x},\bm{y})=\int_{\bm{z}} p_G(\bm{y}|\bm{x},\bm{z})p(\bm{x})p(\bm{z}) \mathrm{d}\bm{z}
    \approx \frac{1}{L}\sum_{i}^L p_G(\bm{y}|\bm{x},\bm{z}_i)p(\bm{x}),
    \label{eq:joint-factor-G}
\end{equation}
where $\bm{z}_i \sim p(\bm{z})$. As suggested in~\cite{Kingma2014}, the number of samples $L$ can be set as 1
as long as the minibatch size is large enough. Under such setting, the pseudo noisy image pair $(\bm{x},\hat{\bm{y}})$ obtained from the generation process in
Eq.~\eqref{eq:sampling-G} can be roughly regarded as an sampled example from $p_G(\bm{x},\bm{y})$.

\subsection{Dual Adversarial Model}

In the previous subsection, we have derived two pseudo joint distributions  
from the perspectives of noise removal and noise generation, i.e., $p_R(\bm{x},\bm{y})$ and $p_G(\bm{x},\bm{y})$.
Now the problem becomes how to effectively train the denoiser $R$ and the generator $G$, in order to
well approximate the joint distribution $p(\bm{x},\bm{y})$. Fortunately, the
tractability of sampling process defined in Eqs.~\eqref{eq:sampling-D} and \eqref{eq:sampling-G}
makes such training possible in an adversarial manner as GAN~\cite{goodfellow2014generative}, which
gradually pushes $p_R(\bm{x},\bm{y})$ and $p_G(\bm{x},\bm{y})$ toward
the true distribution $p(\bm{x},\bm{y})$.
Specifically, we formulate this idea as the following dual adversarial problem inspired by
Triple-GAN~\cite{chongxuan2017triple},
\begin{align}
    \min_{R,G}\max_{D} \mathcal{L}_{\text{gan}}(R,G,D) = E_{(\bm{x},\bm{y})}[D(\bm{x},\bm{y})]
    &- \alpha E_{(\hat{\bm{x}},\bm{y})}[D(\hat{\bm{x}},\bm{y})] \notag \\
    &- (1-\alpha) E_{(\bm{x},\hat{\bm{y}})}[D(\bm{x},\hat{\bm{y}})],
    \label{eq:minmax-pureGan}
\end{align}
where $\hat{\bm{x}}=R(\bm{y})$, $\hat{\bm{y}}=G(\bm{x},\bm{z})$, and $D$ denotes the discriminator,
which tries to distinguish the real clean-noisy image pair $(\bm{x},\bm{y})$ from the fake ones
$(\hat{\bm{x}},\bm{y})$ and $(\bm{x},\hat{\bm{y}})$.
The hyper-parameter $\alpha$ controls the relative importance between the denoiser
$R$ and generator $G$. As in~\cite{arjovsky2017wasserstein}, we use the Wassertein-1 distance to measure
the difference between two distributions in Eq.~\eqref{eq:minmax-pureGan}.

The working mechanism of our dual adversarial network can be intuitively explained in
Fig.~\ref{fig:model-framework}. On one hand, the denoiser $R$, delivering the knowledge of $p_R(\bm{x}|\bm{y})$,
is expected to conduct the joint distribution $p_R(\bm{x},\bm{y})$ of Eq.~\eqref{eq:joint-factor-D},
while the noise generator $G$, conveying the information
of $p_G(\bm{y}|\bm{x},\bm{z})$, is expected to derive the joint distribution $p_G(\bm{x},\bm{y})$ of
Eq.~\eqref{eq:joint-factor-G}.
Through the adversarial effect of discriminator $D$,
the denoiser $R$ and generator $G$ are both gradually optimized so as to pull $p_R(\bm{x},\bm{y})$ and $p_G(\bm{x},\bm{y})$ toward
the true joint distribution $p(\bm{x}, \bm{y})$ during training. On the other hand, the capabilities of $R$ and $G$
are mutually enhanced by their dual regularization between each other. Given any real image pair $(\bm{x}, \bm{y})$
and one pseudo image pair $(\bm{x},\hat{\bm{y}})$ from generator $G$ or $(\hat{\bm{x}}, \bm{y})$ from denoiser $R$,
the discriminator $D$ will be updated according to the adversarial loss. Then $D$ is fixed as a criterion to update
both $R$ and $G$ simultaneously as illustrated by the dotted lines in Fig.~\ref{fig:model-framework}, which means
$R$ and $G$ are keeping interactive and guided by each other in each iteration.

Previous researches~\cite{isola2017image,zhu2017unpaired} have shown that it is benefical to mix the adversarial
objective with traditional losses, which would speed up and stabilize the training of GAN.
For noise removal task, we adopt the $L_1$ loss, i.e., $||\hat{\bm{x}}-\bm{x}||_1$, which enforces
the output of denoiser $R$ to be close to the groundtruth. For the generator $G$, however, the direct $L_1$ loss
would not be benefical because of the randomness of noise.
Therefore, we propose to apply the $L_1$ constrain on the statistical features of noise distribution:
\begin{equation}
    ||\mathcal{GF}(\hat{\bm{y}}-\bm{x}) - \mathcal{GF}(\bm{y}-\bm{x})||_1,
    \label{eq:L1_gauss}
\end{equation}
where $\mathcal{GF}(\cdot)$ represents the Gaussian filter used to extract the first-order statistical
information of noise. Intergrating these two regularizers into the adversarial loss of
Eq.~\eqref{eq:minmax-pureGan}, we obtain the final objective:
\begin{equation}
    \min_{R,G}\max_{D} \mathcal{L}_{gan}(R,G,D)
     + \tau_1 ||\hat{\bm{x}}-\bm{x}||_1 
    + \tau_2 ||\mathcal{GF}(\hat{\bm{y}}-\bm{x}) - \mathcal{GF}(\bm{y}-\bm{x})||_1,
    \label{eq:lossall-pureGan}
\end{equation}
where $\tau_1$ and $\tau_2$ are hyper-parameters to balance different losses.
More sensetiveness analysis on them are provided in Sec.~\ref{sec:SIDD}.

\subsection{Training Strategy} \label{subsec:training}
In the dual adversarial model of Eq.~\eqref{eq:lossall-pureGan}, we have three objects to be
optimized, i.e., the denoiser $R$, generator $G$ and discriminator $D$. As in most of the GAN-related
papers~\cite{goodfellow2014generative,arjovsky2017wasserstein,chongxuan2017triple}, we jointly train
$R$, $G$ and $D$ but update them in an alternating manner as shown in Algorithm~\ref{alg:bianet-joint}.
In order to stabilize the training, we adopt the gradient penalty technology in 
WGAN-GP~\cite{gulrajani2017improved}, enforcing the discriminator to satisfy 1-Lipschitz
constraint by an extra gradient penalty term.

After training, the generator $G$ is able to simulate more noisy images
given any clean images, which are easily obtained from the original training data set or by downloading
from internet. Then we can retrain the denoiser $R$
by adding more synthetic clean-noisy image pairs generated by $G$ to the
training data set. As shown in Sec.~\ref{sec:Experiments}, this strategy can further improve the denoising performance.

\subsection{Network Architecture}
The denoiser $R$, generator $G$ and discriminator $D$ in our framework are all parameterized 
as deep neural networks due to their powerful fitting capability.
As shown in Fig.~\ref{fig:model-framework}, the denoiser $R$ takes noisy image $\bm{y}$ as input and outputs denoised
image $\hat{\bm{x}}$, while the generator $G$ takes the concatenated clean image $\bm{x}$ and latent variable
$\bm{z}$ as input and outputs the
simulated noisy image $\hat{\bm{y}}$. For both $R$ and $G$, we use the UNet~\cite{Ronneberger2015} architecture as
backbones. Besides, the residual learning strategy~\cite{zhang2017beyond} is adopted in both of them.
The discriminator $D$ contains
five stride convolutional layers to reduce the image size and one fully connected layer to fuse all the information.
More details about the network architectures are provided in the supplementary material due to
page limitation. It should be noted that our proposed method is a general framework that does not depend on the
specific architecture, therefore most of the commonly used networks architectures~\cite{zhang2017beyond,Mao2016,anwar2019real}
in low-level vision tasks can be substituted.

\begin{algorithm}[t]
    \caption{Daul adversarial network.} \label{alg:bianet-joint}
    \begin{algorithmic}[1]
        \REQUIRE hyper-parameters: $\tau_1$, $\tau_2$, $\alpha$, $n_{critic}$
        \WHILE {$\theta$ had not converged}
        \FOR {$i=1,2, \dots, n_{critic}$}
        \STATE Sample a batch of pairs $(\bm{x},\bm{y})$ from $p(\bm{x},\bm{y})$
        \STATE Sample a batch of pairs $(\hat{\bm{x}},\bm{y})$ from $p_R(\bm{x},\bm{y})$
               and $(\bm{x},\hat{\bm{y}})$ from $p_G(\bm{x},\bm{y})$ 
        \STATE Update discriminator $D$ with fixed $R$ and $G$
        \ENDFOR
        \STATE Update denoiser $R$ with fixed $G$ and $D$
        \STATE Update generator $G$ with fixed $R$ and $D$
        \ENDWHILE
    \end{algorithmic}
\end{algorithm}
\vspace{-2mm}

\section{Evaluation Metrics}\label{sec:metric}

For the noise removal task, PSNR and SSIM~\cite{wang2004image} can be readily adopted to compare the denoising
performance of different methods.
However, to the best of our knowledge, there is still no any quantitative metric having been designed for noise
generation task.
To address this issue, we propose two metrics to compare the similarity between
the generated and the real noisy images as follows:
\begin{itemize}
    \item \textbf{PGap} (PSNR Gap): The main idea of PGap is to compare the synthetic and real noisy images indirectly
        by the performance of the denoisers trained on them. Let $\mathcal{D}=\{(\bm{x}^i, \bm{y}^i)\}_{i=1}^N$,
        $\mathcal{T}=\{(\tilde{\bm{x}}^j, \tilde{\bm{y}}^j)\}_{j=1}^S$ denote the available training and testing sets, whose
        noise distributions are same or similar. Given any one noisy image generator $G$, we can 
        synthesize another training set:
        \begin{small}
        \begin{equation}
            \mathcal{D}_G=\{(\bm{x}^i, \tilde{\bm{y}}^i)|\tilde{\bm{y}}^i=G(\bm{x}^i,\bm{z}^i),
            \bm{z}^i\sim p(\bm{z})\}_{i=1}^N.
            \label{eq:datatet-G}
        \end{equation}
        \end{small}
        \hspace{-1mm}After training two denoisers $R_1$ on the original
        data set $\mathcal{D}$ and 
        $R_2$ on the generated data set $\mathcal{D}_{G}$ under the \textbf{same }conditions, we can define PGap as
        \begin{small}
        \begin{equation}
            \text{PGap} = \text{PSNR}(R_1(\mathcal{T})) - \text{PSNR}(R_2(\mathcal{T})),
            \label{eq:PGap_definition}
        \end{equation}
        \end{small}
        \hspace{-1mm}where $\text{PSNR}(R_i(\mathcal{T})) (i=1, 2)$ represents the PSNR result of denoiser $R_i$ on
        testing data set $\mathcal{T}$. It is obvious that, if the generated noisy images in $\mathcal{D}_G$
        are close to the real noisy ones in $\mathcal{D}$, the performance of $R_2$ would be close to
        $R_1$, and thus the PGap would be small.
        \vspace{2mm}
    \item \textbf{AKLD} (Average KL  Divergence): The noise generation task aims at synthesizing fake noisy image
        $\bm{y}^f$ from the real clean image $\bm{x}^r$ to match the real noisy image $\bm{y}^r$ in distribution.
        Therefore, the KL divergence between the conditional distributions $p_{\bm{y}^f}(\bm{y}|\bm{x})$ on the fake
        image pair $(\bm{x}^r,\bm{y}^f)$ and $p_{\bm{y}^r}(\bm{y}|\bm{x})$ on the real image pair $(\bm{x}^r, \bm{y}^r)$
        can serve as a metric. To make this conditional distribution tractable, we utlize the pixel-wisely Gaussian
        assumption for real noise in recent work VDN~\cite{yue2019variational}, i.e.,
        \begin{small}
        \begin{shrinkeq}{-1mm}{-1mm}
        \begin{equation}
            p_{\bm{y}^c}(\bm{y}|\bm{x}) = \mathcal{N}(\bm{y}|[\bm{x}^r], \text{diag}([\bm{V}^c])), ~ c \in \{f,r\},
            \label{eq:VDN-Distribution}
        \end{equation}
        \end{shrinkeq}
        \end{small}
        \hspace{-1mm}where
        \begin{small}
        \begin{equation}
            \bm{V}^c = \mathcal{GF}((\bm{y}^c - \bm{x}^r)^2), ~ c \in \{f, r\},
            \label{eq:sigma-estimate}
        \end{equation}
        \end{small}
        \hspace{-1mm}$[\cdot]$ denotes the reshape operation from matrix to vector, $\mathcal{GF}(\cdot)$ denotes the Gaussian filter, 
        and the square of $(\bm{y}^c-\bm{x}^r)^2$ is pixel-wise operation.
        Based on such explicit distribution assumption, the KL divergence between $p_{\bm{y}^f}(\bm{y}|\bm{x})$ and
        $p_{\bm{y}^r}(\bm{y}|\bm{x})$ can be regarded as an intuitive metric. To reduce the influence of randomness,
        we randomly generate $L$ synthetic fake noisy images:
        \begin{small}
        \begin{equation}
            \bm{y}^{f_j} = G(\bm{x}^r, \bm{z}^j), ~ \bm{z}^j \sim p(\bm{z}), ~j=1,2,\cdots,L,
            \label{eq:yj_generate}
        \end{equation}
        \end{small}
        \hspace{-1mm}for any real clean image $\bm{x}^r$, and define the following average KL divergence as our metric, i.e.,
        \begin{small}
        \begin{shrinkeq}{-1mm}{-2mm}
        \begin{equation}
            \text{AKLD} = \frac{1}{L}\sum_{j=1}^L KL[p_{\bm{y}^{f_j}}p(\bm{y}|\bm{x})||p_{\bm{y}^r}(\bm{y}|\bm{x})].
            \label{eq:average-KL}
        \end{equation}
        \end{shrinkeq}
        \end{small}
        \hspace{-1mm}Evidently, the smaller AKLD is, the better the generator $G$ is. In the following experiments, we set 
        $L=50$.
\end{itemize}

\section{Experimental Results} \label{sec:Experiments}

In this section, we conducted a series of experiments on several real-world denoising benchmarks.
In specific, we considered two groups of experiments:
the first group (Sec.~\ref{sec:SIDD}) is designed for evaluating
the effectiveness of our method on both of the noise removal and noise generation tasks, which is implemented
on one specific real benchmark containing training, validation and testing sets; while the second
group (Sec.~\ref{sec:other-benchmark}) is conducted on two real benchmarks that only consist of some noisy
images as testing set, aiming at evaluating its performance on general real-world denoising tasks.
Due to the page limitation, the running time comparisons are listed in the supplementary material.

In brief, we denote the jointly trained \textbf{D}ual \textbf{A}dversarial \textbf{Net}work following
Algorithm~\ref{alg:bianet-joint} as DANet. As discussed in Sec.~\ref{subsec:training}, the learned
generator $G$ in DANet is able to augment the original
training set by generating more synthetic clean-noisy image pairs, and the retrained denoiser $R$ on this 
augmented training data set under $L_1$ loss is denoted as $\text{DANet}_{+}$.

\subsection{Experimental Settings}

\textbf{Parameter settings and network training}: In the training stage of DANet, the weights of $R$ and
$G$ were both initialized according to~\cite{he2015delving}, and the weights of $D$ were
initialized from a zero-centered Normal
distribution with standard deviation 0.02 as~\cite{Radford2016}. All the three networks were trained by
Adam optimizer~\cite{Kingma2015} with momentum terms $(0.9,0.999)$ for $R$ and $(0.5, 0.9)$
for both $G$ and $D$. The learning rates were set as $1e\text{-}4$, $1e\text{-}4$ and $2e\text{-}4$ for $R$, $G$ and
$D$, respectively, and linearly decayed in half every 10 epochs.

In each epoch, we randomly cropped $16\times 5000$ patches with size $128\times 128$ from the images for training.
During training, we updated $D$ three times for each update of $R$ and $G$.
We set $\tau_1=1000$, $\tau_2=10$ throughout
the experiments, and the sensetiveness analysis about them can be found in
Sec.~\ref{sec:SIDD}. As for $\alpha$, we set it as $0.5$, meaning the denoiser $R$ and 
generator $G$ contribute equally in our model. The penalty coefficient in
WGAN-GP~\cite{gulrajani2017improved} is set as 10 following its default settings. 
As for $\text{DANet}_{+}$, the denoiser $R$ was retrained with the same settings as that in DANet.  
All the models were trained using PyTorch~\cite{paszke2019pytorch}.

\begin{table}[t]
\begin{flushleft}
\makeatletter\def\@captype{table}\makeatother
\begin{minipage}[c][3.6cm][c]{0.58\textwidth}
    \caption{The PGap and AKLD performances of different compared methods on the SIDD validation
        data set. And the best results are highlighted in bold.}
    \small
    \begin{tabular}{@{}C{1.5cm}@{}|@{}C{1.6cm}@{}@{}C{1.2cm}@{}@{}C{1.3cm}@{}@{}C{1.4cm}@{}}
        \Xhline{0.8pt}
        \multirow{2}*{Metrics}   & \multicolumn{4}{c}{Methods} \\
        \Xcline{2-5}{0.4pt}
                            & CBDNet & ULRD   & GRDN   & DANet         \\
        \Xhline{0.4pt}
        PGap$\downarrow$    & 8.30   & 4.90   & 2.28   & \textbf{2.06}  \\
        \Xhline{0.4pt}
        AKLD$\downarrow$    & 0.728  & 0.545  & 0.443  & \textbf{0.212}  \\
        \Xhline{0.8pt}
    \end{tabular}
    \label{tab:PGap}
\end{minipage} %
\hspace{5mm}
\makeatletter\def\@captype{figure}\makeatother
\begin{minipage}[c][3.6cm][c]{0.34\textwidth}
    \centering
    \vspace{4mm}
    \includegraphics[keepaspectratio, height=2.8cm]{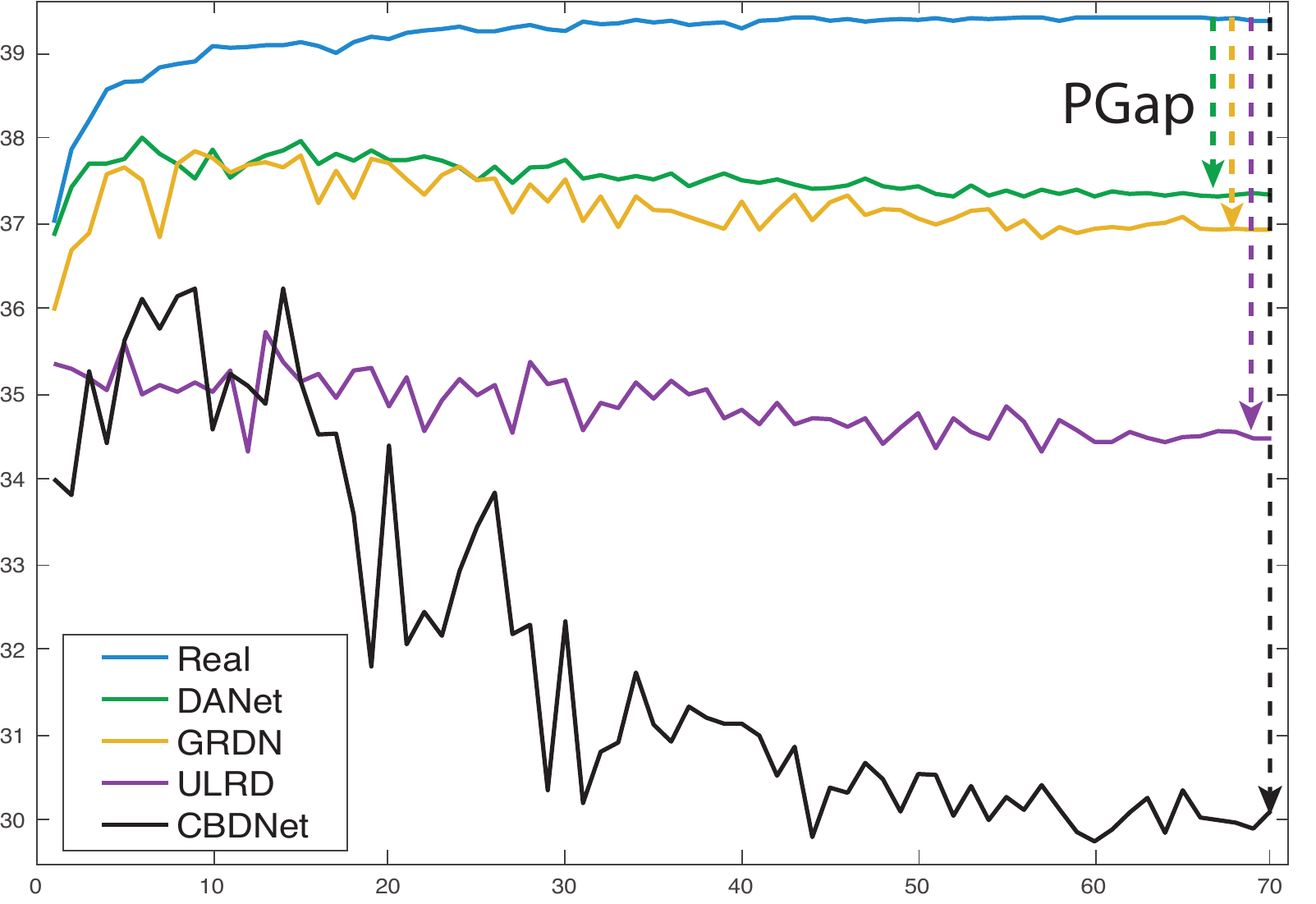}
    \caption{\scriptsize PSNR results of different methods during training.}
    \label{fig:PGap}
\end{minipage}
\end{flushleft}
\vspace{-6mm}
\end{table}
\subsection{Results on SIDD Benchmark} \label{sec:SIDD}
In this part, SIDD~\cite{SIDD_2018_CVPR} benchmark
is employed
to evaluate the denoising performance and generation quality of our proposed method. The full SIDD data set contains
about $24000$ clean-noisy image pairs as training data, and the rest $6000$ image pairs are held as the benchmark
for testing. For fast training and evaluation, one medium training set (320 image pairs) and validation
set (40 image pairs) are also provided, but the testing results can only be obtained by 
submission. We trained DANet and $\text{DANet}_{+}$ on the medium version training set,
and evaluated on the validation and testing sets.

\begin{figure}[t]
    \centering
    \includegraphics[scale=0.55]{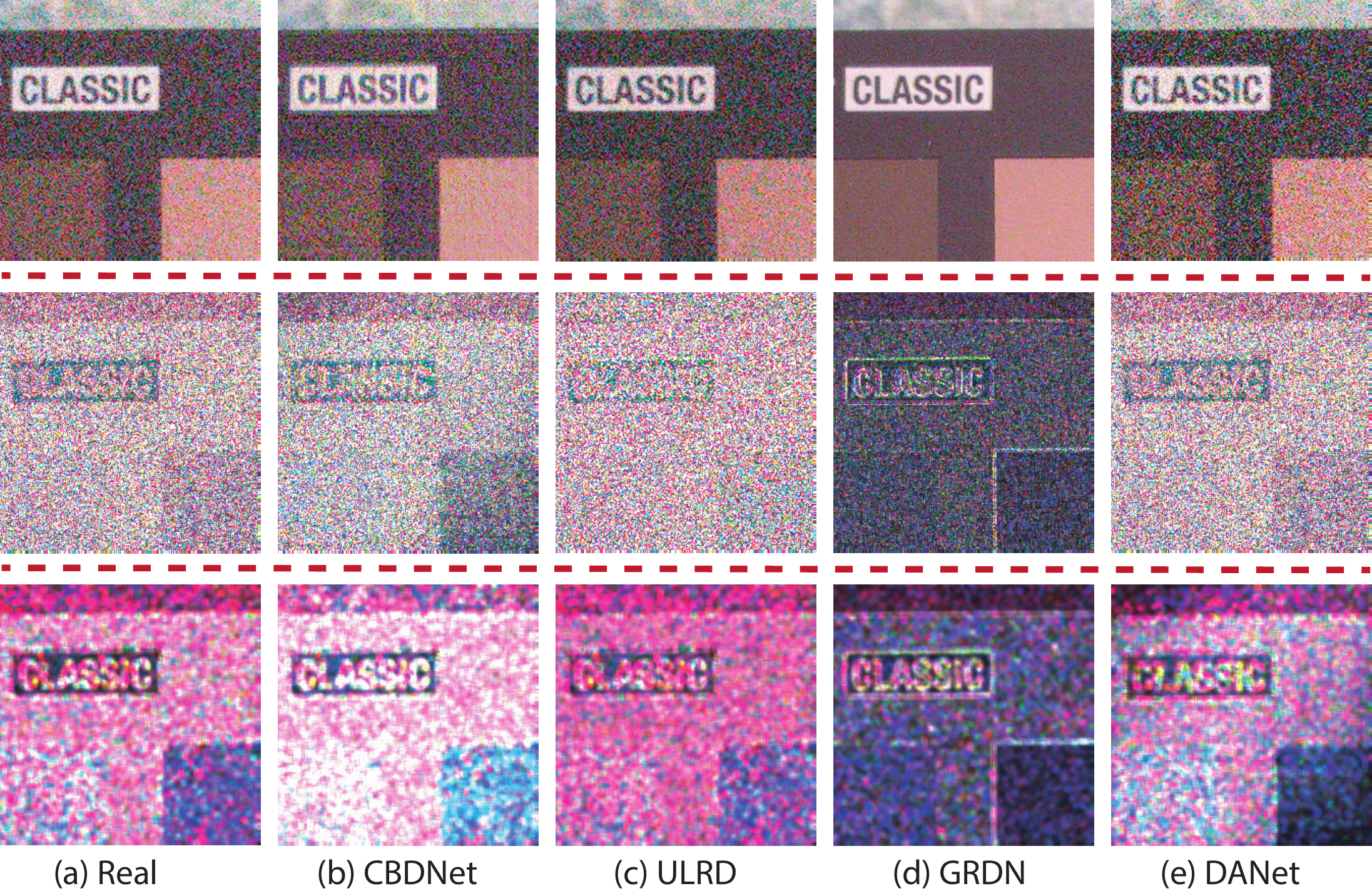}
    \vspace{-7mm}
    \caption{Illustration of one typical generated noisy images (1st row) by different methods and their corresponding
        noise (2nd row) and variance map (3rd row) estimated by Eq.~\eqref{eq:sigma-estimate}. The first
        column represents the real ones in SIDD validation set.
    } \label{fig:generation-compare}
\end{figure}
\vspace{2mm}\noindent\textbf{Noise Generation:} 
The generator $G$ in DANet is mainly used to synthesize the corresponding noisy image given any
clean one. As introduced in Sec.~\ref{sec:metric},
two metrics PGap and AKLD are designed to assess the generated noisy image.
Based on these two metrics, we compared DANet with three recent methods, including CBDNet~\cite{guo2019toward},
ULRD~\cite{brooks2019unprocessing} and GRDN~\cite{kim2019grdn}. CBDNet and ULRD both attempted to generate noisy
images by simulating the in-camera processing pipelines, while GRDN directly learned the noise distribution
using GAN~\cite{goodfellow2014generative}. It should be noted that ULRD~\cite{brooks2019unprocessing} and
GRDN~\cite{kim2019grdn} both make use of the metadata of the images.

Table~\ref{tab:PGap} lists the PGap values of different compared methods on SIDD validation set.
For the calculation of PGap, SIDD validation set is regarded as the testing set $\mathcal{T}$ in
Eq.~\eqref{eq:PGap_definition}. Obviously, our proposed DANet achieves the best performance. Figure~\ref{fig:PGap}
displays the PSNR curves of different denoisers trained on the real training set or only the synthetic
training sets generated by different methods, which gives an intuitive illustration for our defined
PGap. It can be seen that all the methods tend to gradually overfit to their own
synthetic training set, especially for CBDNet.
However, DANet performs not only more stably but also better than other methods.
\begin{table}[t]
    \centering
    \caption{The PSNR and SSIM results of different methods on SIDD validation and testing sets. The best results
    are highlighted in bold.} \label{tab:SIDD-val-test}
    \vspace{-3mm}
    \scriptsize
    \begin{tabular}{@{}C{1.4cm}@{}|c|@{}C{1.4cm}@{}@{}C{1.1cm}@{}@{}C{1.2cm}@{}@{}C{1.2cm}@{}@{}C{1.2cm}@{}
        @{}C{1.0cm}@{}@{}C{1.2cm}@{}@{}C{1.3cm}@{}}
        \Xhline{0.8pt}
        \multirow{2}*{Datasets} & \multirow{2}*{Metrics} & \multicolumn{8}{c}{Methods} \\
        \Xcline{3-10}{0.4pt}
        &   & CBM3D  & WNNM & DnCNN & CBDNet & RIDNet & VDN & DANet & $\text{DANet}_{+}$  \\
        \Xhline{0.4pt}
        \multirow{2}*{Testing} & PSNR$\uparrow$ & 25.65 & 25.78 & 23.66 &33.28 &- & 39.26 &39.25 &\textbf{39.43} \\
        \Xcline{2-10}{0.4pt}
                               & SSIM$\uparrow$ & 0.685 & 0.809 & 0.583 &0.868 &- & 0.955 &0.955 &\textbf{0.956} \\
        \Xhline{0.4pt}
        \multirow{2}*{Validation} & PSNR$\uparrow$ & 25.29 & 26.31 & 38.56 &38.68 &38.71 & 39.29 & 39.30 & \textbf{39.47} \\
        \Xcline{2-10}{0.4pt}
                                  & SSIM$\uparrow$ & 0.412 & 0.524 & 0.910 &0.909 &0.913 & 0.911 & 0.916 & \textbf{0.918} \\
        \Xhline{0.8pt}
    \end{tabular}
    \vspace{-3mm}
\end{table}

The average AKLD results calculated on all the images of SIDD validation set are also listed in Table~\ref{tab:PGap}.
The smallest AKLD of DANet indicates that it learns a better implicit distribution to
approximate the true distribution $p(\bm{y}|\bm{x})$.
Fig.~\ref{fig:generation-compare} illustrates one typical example of the real and synthetic noisy images generated by
different methods, which provides an intuitive visualization for the AKLD metric. In summary, DANet
outperforms other methods both in quantization and visualization, even though some of them make use of additional metadata.

\begin{figure}[t]
    \centering
    \includegraphics[scale=0.55]{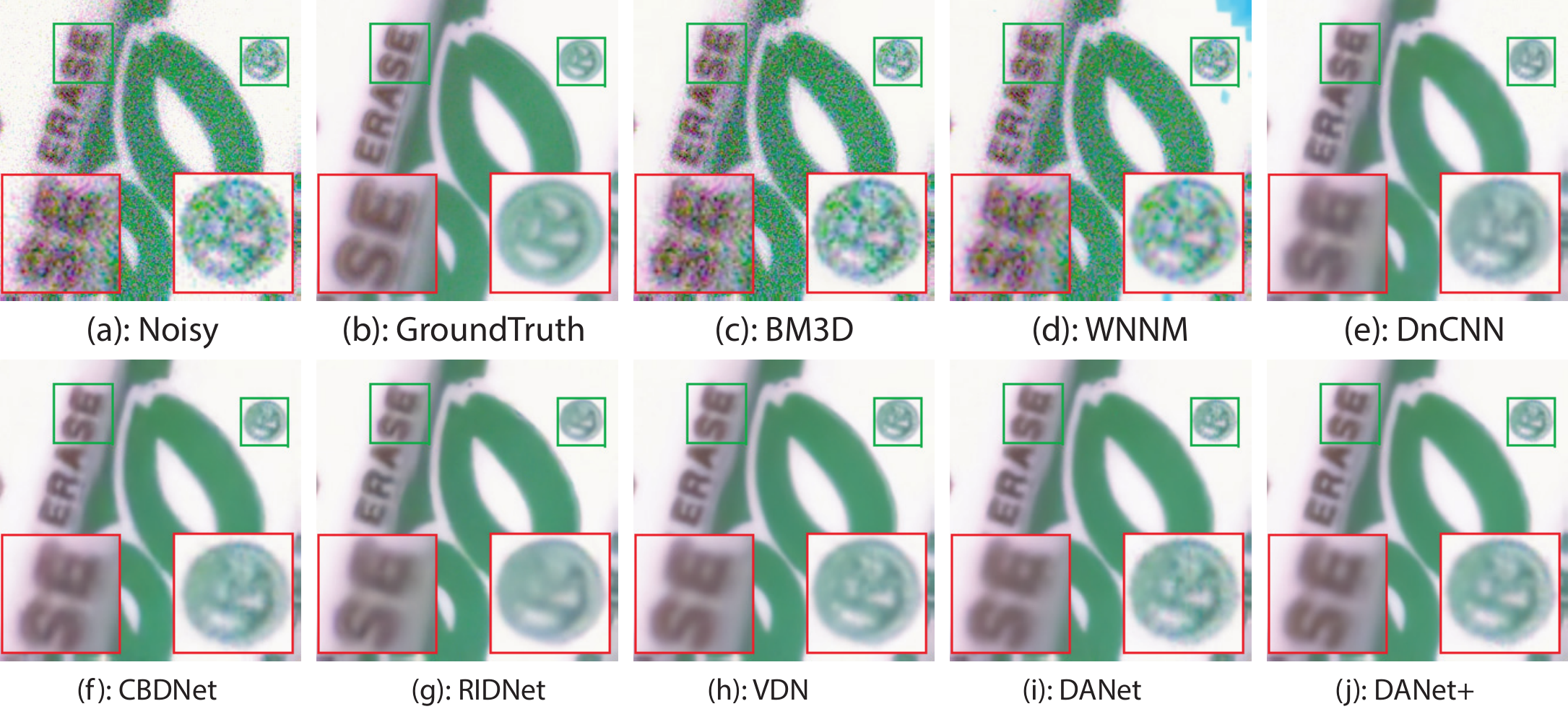}
    \vspace{-6mm}
    \caption{One typical denoising example in the SIDD validation
    dataset.}
    \label{fig:SIDD-val_denoising1}
\end{figure}
\begin{table}[t]
    \parbox[b]{.45\textwidth}{ \centering
    \caption{The PSNR and SSIM results of DANet under different $\tau_1$ values
    on SIDD validation data set.} \label{tab:tau1}
    \vspace{-3mm}
    \small
    \begin{tabular}{@{}C{1.2cm}@{}|@{}C{1.05cm}@{}@{}C{1.05cm}@{}@{}C{1.05cm}@{}@{}C{1.05cm}@{}}
        \Xhline{0.8pt}
        \multirow{2}*{Metrics}   & \multicolumn{4}{c}{$\tau_1$} \\
        \Xcline{2-5}{0.4pt}
                           & $1e\text{+}2$  & $1e\text{+}3$  & $1e\text{+}4$    & +$\infty$ \\
        \Xhline{0.4pt}
        PSNR$\uparrow$     & 38.66          & 39.30          & 39.33            & 39.39  \\
        \Xhline{0.4pt}
        SSIM$\uparrow$     & 0.901          & 0.916          & 0.916            & 0.917  \\
        \Xhline{0.8pt}
    \end{tabular}} %
    \hspace{1mm}
    \parbox[b]{.51\textwidth}{ \raggedleft
    \caption{The PGap and AKLD results of DANet under different $\tau_2$ values
    on SIDD validation data set.} \label{tab:tau2}
    \vspace{-3mm}
    \small
    \begin{tabular}{@{}C{1.2cm}@{}|@{}C{1.0cm}@{}@{}C{1.0cm}@{}@{}C{1.0cm}@{}@{}C{1.0cm}@{}@{}C{1.0cm}@{}}
        \Xhline{0.8pt}
        \multirow{2}*{Metrics}   & \multicolumn{5}{c}{$\tau_2$} \\
        \Xcline{2-6}{0.4pt}
                            & 0       & 5      & 10    & 50     & +$\infty$\\
        \Xhline{0.4pt}
        PGap$\downarrow$    & 5.33    & 3.10   & 2.06  & 4.17   & 15.14  \\
        \Xhline{0.4pt}
        AKLD$\downarrow$    & 0.386   & 0.216  & 0.212 & 0.177  & 0.514  \\
        \Xhline{0.8pt}
    \end{tabular}}
\end{table}

\vspace{2mm}\noindent\textbf{Noise Removal:}
To verify the effectiveness of our proposed method on real-world denoising task, we compared it
with several state-of-the-art methods, including CBM3D~\cite{dabov2007image}, WNNM~\cite{gu2014weighted},
DnCNN~\cite{zhang2017beyond}, CBDNet~\cite{guo2019toward}, RIDNet~\cite{anwar2019real} and
VDN~\cite{yue2019variational}. Table~\ref{tab:SIDD-val-test}
lists the PSNR and SSIM results of different methods on SIDD validation and testing sets. It should be noted that
the results on testing sets are cited from official
website\footnote{\url{https://www.eecs.yorku.ca/~kamel/sidd/benchmark.php}}, but the results on validation set are calculated
by ourself. For fair comparison, we retrained DnCNN and CBDNet on SIDD training set.
From Table~\ref{tab:SIDD-val-test},
it is easily observed that: 1) deep learning methods obviously performs better than traditional
methods CBM3D and WNNM due to the powerful fitting capability
of DNN; 2) DANet and $\text{DANet}_{+}$ both outperform the state-of-the-art real-world denoising
methods, substantiating their effectiveness;
3) $\text{DANet}_{+}$ surpasses DANet about 0.18dB PSNR, which indicates that the synthetic data by $G$ facilitates
the training of the denoiser $R$.

Fig.~\ref{fig:SIDD-val_denoising1} illustrates the visual denoising results of different methods. It can be seen that
CBM3D and WNNM both fail to remove the real-world noise. DnCNN tends to produce over-smooth edges and textures
due to the $L_2$ loss. CBDNet, RIDNet and VDN alleviate this phenomenon to some extent since they adopt more robust loss
functions.
DANet recovers sharper edges and more details owning to the
adversarial loss.
After retraining with
more generated image pairs, $\text{DANet}_{+}$ obtains the closer denoising results to the groundtruth.
\begin{figure}[t]
    \centering
    \includegraphics[scale=0.69]{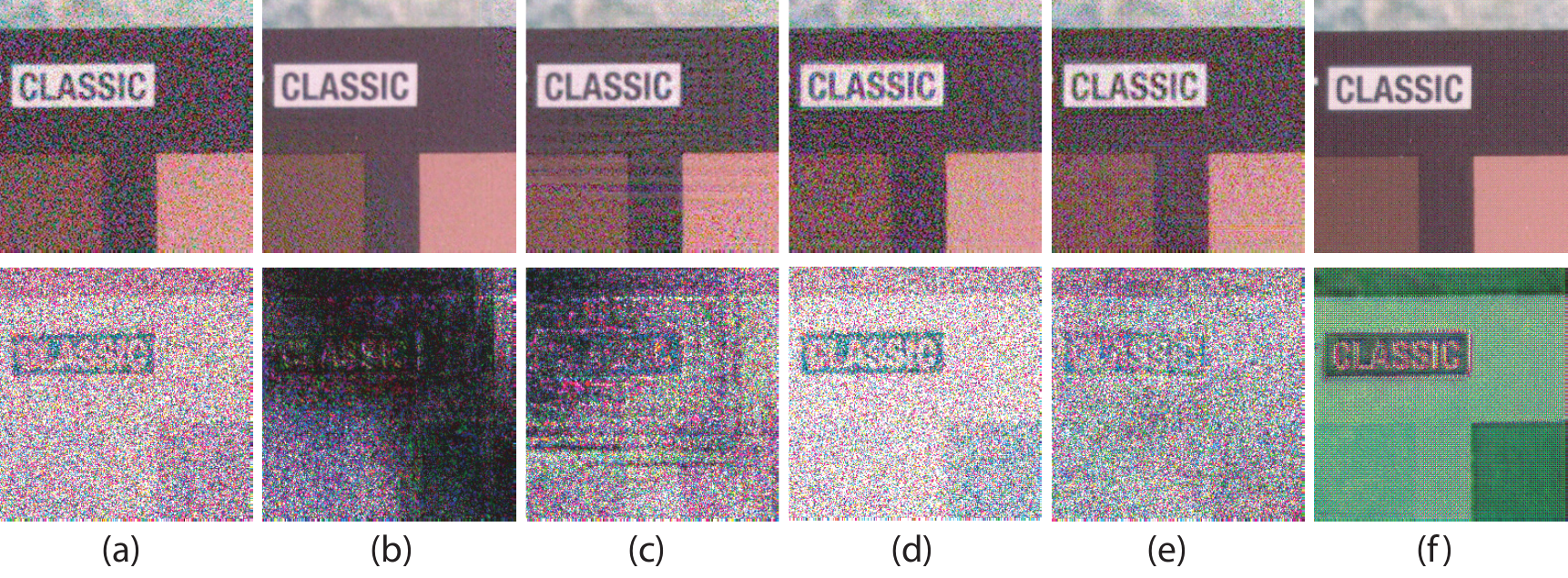}
    \vspace{-3mm}
    \caption{This figure displays the real or generated noisy images (the 1st row) by DANet under different $\tau_2$
    value and the corresponding noise (the 2nd row). From left to right: (a) real case, (b) $\tau_2=0$, (c) $\tau_2=5$,
    (d) $\tau_2=10$, (e) $\tau_2=50$, (f) $\tau_2=+\infty$.}
    \label{fig:generate-tau}
\end{figure}

\vspace{2mm}\noindent\textbf{Hyper-parameter Analysis:} 
Our proposed DANet involves two hyper-parameters $\tau_1$ and $\tau_2$ in Eq.~\eqref{eq:lossall-pureGan}. The
pamameter $\tau_1$ mainly influences the performance of denoiser $R$, while $\tau_2$ directly affects the
generator $G$.

\begin{table}[t]
    \parbox[b]{.49\textwidth}{ \centering
    \caption{The comparison results of BaseD and DANet on SIDD validation set.} \label{tab:BaseD}
    \vspace{-3mm}
    \small
    \begin{tabular}{@{}C{1.9cm}@{}|@{}C{2.0cm}@{}@{}C{2.0cm}@{}}
        \Xhline{0.8pt}
        \multirow{2}*{Metrics}   & \multicolumn{2}{c}{Methods} \\
        \Xcline{2-3}{0.4pt}
                           & BaseD          & DANet  \\
        \Xhline{0.4pt}
        PSNR$\uparrow$     & 39.19          & 39.30          \\
        \Xhline{0.4pt}
        SSIM$\uparrow$     & 0.907          & 0.916          \\
        \Xhline{0.8pt}
    \end{tabular}} %
    \hspace{1mm}
    \parbox[b]{.49\textwidth}{ \raggedleft
    \caption{The comparison results of BaseG and DANet on SIDD validation set.} \label{tab:BaseG}
    \vspace{-3mm}
    \small
    \begin{tabular}{@{}C{2.0cm}@{}|@{}C{2.0cm}@{}@{}C{2.0cm}@{}}
        \Xhline{0.8pt}
        \multirow{2}*{Metrics}   & \multicolumn{2}{c}{Methods} \\
        \Xcline{2-3}{0.4pt}
                            & BaseG       & DANet       \\
        \Xhline{0.4pt}
        PGap$\downarrow$    & 4.07        & 2.06    \\
        \Xhline{0.4pt}
        AKLD$\downarrow$    & 0.223       & 0.212   \\
        \Xhline{0.8pt}
    \end{tabular}}
\end{table}
Table~\ref{tab:tau1} lists the PSNR/SSIM results of DANet under different $\tau_1$ settings, where $\tau_1=+\infty$
represents the results of the denoiser $R$ trained only with $L_1$ loss.
As expected, small $\tau_1$ value, meaning that the adversarial loss
plays more important role, leads to the decrease of PSNR and SSIM performance
to some extent. However, when $\tau_1$ value is too large, the $L_1$ regularizer will mainly dominates the
performance of denoiser $R$. Therefore, we set $\tau_1$ as a moderate value $1e\text{+}3$ throughout all the
experiments, which makes the denoising results more realistic as shown in Fig.~\ref{fig:SIDD-val_denoising1}
even sacrificing a little PSNR performance.

The PGap and average AKLD results of DANet under different $\tau_2$ values are shown in Table~\ref{tab:tau2}. Note that
$\tau_2=+\infty$ represents the results of the generator $G$ trained only with the regularizer of
Eq.~\eqref{eq:L1_gauss}. Fig.~\ref{fig:generate-tau} also shows
the corresponding visual results of one typical example. As one can see, $G$ fails to simulate the real noise
with $\tau_2=0$, which demonstrates that the regularizer of
Eq.~\eqref{eq:L1_gauss} is able to stabilize the training of GAN. However, it is also difficult to train $G$
only with the regularizer of Eq.~\eqref{eq:L1_gauss} as shown in Fig.~\ref{fig:generate-tau} (f). Taking both
the quantitative and visual results into consideration, $\tau_2$ is constantly set as $10$ in our experiments.

\vspace{2mm}\noindent\textbf{Ablation studies:} To verify the marginal benefits brought up by our dual adversarial loss,
two groups of ablation experiments are designed in this part. In the first group, we train DANet without the generator and denote
the trained model as BaseD. On the contrary, we train DANet without the denoiser and denote the trained model as BaseG. 
And the comparison results of these two baselines with DANet on noise removal and noise generation tasks are listed in Table~\ref{tab:BaseD}  and
Table~\ref{tab:BaseG}, respectively. It can be easily seen that DANet achieves better performance than both the two baselines in
noise removal and noise generation tasks, especially in the latter, which illustrates the mutual guidance and amelioration between
the denoiser and the generator.
\subsection{Results on DND and Nam Benchmarks}\label{sec:other-benchmark}
To evaluate the performance of our method in general real-world denoising tasks, we test on two real-world
benchmarks, i.e., DND~\cite{plotz2017benchmarking} and Nam~\cite{nam2016holistic}.
These two benchmarks do not provide any training data, therefore they are suitable to test the
generalization capability of any denoiser. Following the experimental setting in RIDNet~\cite{anwar2019real}, we
trained another model using $512\times 512$ image patches from SIDD~\cite{SIDD_2018_CVPR}, Poly~\cite{xu2018real}
and RENOIR~\cite{Anaya2014} for fair comparison. To be distinguished from the model of Sec.~\ref{sec:SIDD}, the
trained models under this setting are denoted as GDANet and $\text{GDANet}_{+}$, aiming at dealing with the
general denoising task in real application. For the training of $\text{GDANet}_{+}$, we employed the images of
MIR Flickr~\cite{Huiskes2010} as clean images to synthesize more training pairs using $G$.

\begin{table}[t]
    \centering
    \caption{The PSNR and SSIM results of different methods on DND benchmark. The best results
    are highlighted as bold.} \label{tab:DND-test}
    \vspace{-3mm}
    \scriptsize
    \begin{tabular}{@{}C{1.3cm}@{}|@{}C{1.3cm}@{}@{}C{1.2cm}@{}@{}C{1.3cm}@{}@{}C{1.4cm}@{}@{}C{1.3cm}@{}
        @{}C{1.1cm}@{}@{}C{1.4cm}@{}@{}C{1.5cm}@{}}
        \Xhline{0.8pt}
        \multirow{2}*{Metrics} & \multicolumn{8}{c}{Methods} \\
        \Xcline{2-9}{0.4pt}
                        & CBM3D  & WNNM   & DnCNN  & CBDNet & RIDNet & VDN    & GDANet & $\text{GDANet}_{+}$  \\
        \Xhline{0.4pt}
         PSNR$\uparrow$ & 34.51  & 34.67  & 32.43  &38.06   &39.26   & 39.38  &39.47    &\textbf{39.58} \\
        \Xhline{0.4pt}
         SSIM$\uparrow$ & 0.8244 & 0.8646 & 0.7900 &0.9421  &0.9528  & 0.9518 &\textbf{0.9548}   &0.9545 \\
         \Xhline{0.8pt}
    \end{tabular}
    \vspace{-1mm}
\end{table}
\begin{figure}[t]
    \centering
    \includegraphics[scale=0.69]{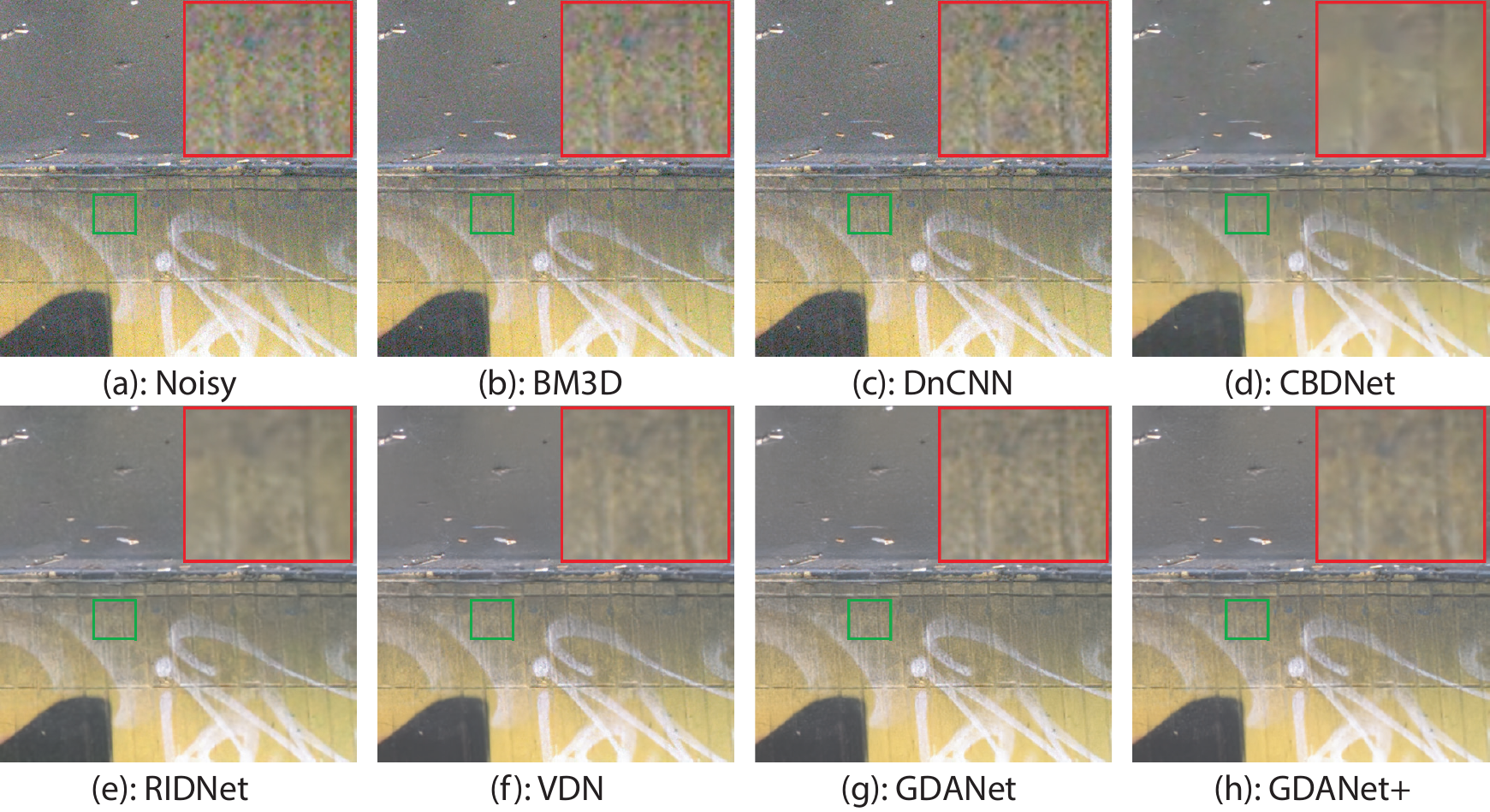}
    \vspace{-6mm}
    \caption{Denoising results of different methods on DND benchmark.} \label{fig:dnd-fig1}
    \vspace{-2mm}
\end{figure}
\begin{table}[t]
    \centering
    \caption{The PSNR and SSIM results of different methods on the Nam benchmark. The best results
    are highlighted in bold.} \label{tab:Nam-test}
    \vspace{-3mm}
    \scriptsize
    \begin{tabular}{@{}C{1.3cm}@{}|@{}C{1.3cm}@{}@{}C{1.2cm}@{}@{}C{1.3cm}@{}@{}C{1.4cm}@{}@{}C{1.3cm}@{}
        @{}C{1.1cm}@{}@{}C{1.4cm}@{}@{}C{1.5cm}@{}}
        \Xhline{0.8pt}
        \multirow{2}*{Metrics} & \multicolumn{8}{c}{Methods} \\
        \Xcline{2-9}{0.4pt}
                        & CBM3D  & WNNM   & DnCNN  & CBDNet & RIDNet & VDN    & GDANet         & $\text{GDANet}_{+}$  \\
        \Xhline{0.4pt}
        PSNR$\uparrow$  & 35.36  & 35.33  & 35.68  &39.20   & 39.33  & 38.66  &\textbf{39.91}  &39.79 \\
        \Xhline{0.4pt}  
        SSIM$\uparrow$  & 0.8708 & 0.8812 & 0.8811 &0.9676  & 0.9623 & 0.9613 &\textbf{0.9693} &0.9689 \\
        \Xhline{0.8pt}
    \end{tabular}
    \vspace{-1mm}
\end{table}
\vspace{2mm}\noindent\textbf{DND Benchmark:}
This benchmark contains 50 real noisy and almost noise-free image pairs.  However, the almost noise-free images
are not publicly released, thus the PSNR/SSIM results can only be obtained through online submission system.
Table~\ref{tab:DND-test} lists the PSNR/SSIM results released on the official DND benchmark
website\footnote{\url{https://noise.visinf.tu-darmstadt.de/benchmark/}}. From Table~\ref{tab:DND-test}, we have the
following observations: 1) $\text{GDANet}_{+}$ outperforms the state-of-the-art VDN about 0.2dB PSNR, which is a large
improvement in the field of real-world denoising; 2) GDANet obtains the highest SSIM value, which
means that it preserves more structural information than other methods as that can be visually observed
in Fig.~\ref{fig:dnd-fig1};
3) DnCNN cannot remove most of the real noise because it overfits to the Gaussian noise case;
4) the classical CBM3D and WNNM methods cannot handle the complex real noise.
\begin{figure}[t]
    \centering
    \includegraphics[scale=0.55]{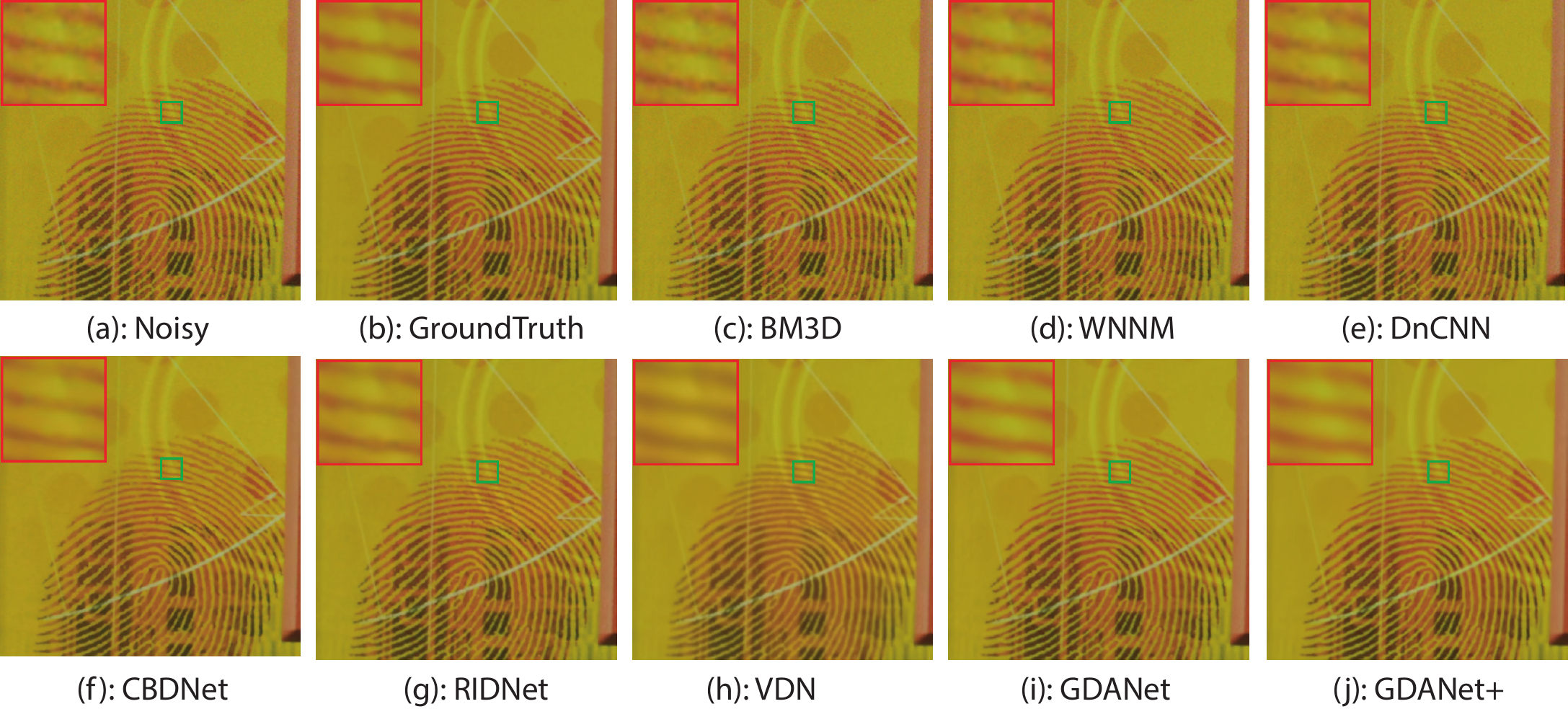}
    \vspace{-6mm}
    \caption{One typical denoising example of Nam benchmark by different methods.} \label{fig:nam-fig1}
\end{figure}

\vspace{2mm}\noindent\textbf{Nam Benchmark:} This benchmark contains 11 real static scenes and the corresponding
noise-free images,
which are obtained by averaging 500 noisy images of the same scenes. We cropped these images into $512\times 512$
patches, and randomly selected 100 of them for the purpose of evaluation.
The quantitative PSNR and SSIM results are given in Table~\ref{tab:Nam-test}. It is easy to see that our proposed GDANet
performs better than the other compared methods.
Note that VDN does not achieve good performance since the noisy images in
this benchmark are JPEG compressed, which is not considered in VDN. For easy comparison, we also display one typical
denoised example by different methods in Fig.~\ref{fig:nam-fig1}, and the better visual performance of our methods
can be observed.

\vspace{2mm}\noindent\textbf{Discussion:} Different from the results in Sec.~\ref{sec:SIDD},
GDANet performs more stably than $\text{GDANet}_{+}$ 
as shown in Table~\ref{tab:DND-test} and \ref{tab:Nam-test}, especially on SSIM metric.
That's because the noise types 
simulated by the generator $G$, which are mainly determined by the training data set, does not match well with that contained in
the testing set. Therefore, GDANet is suggested to be used
in such general real-world denoising task with uncertain noise types, while $\text{DANet}_{+}$ is more suitable in the 
scenario that provides similar training and testing data sets.


\vspace{-2mm}\section{Conclusion}\vspace{-2mm}
We have proposed a new Bayesian framework, namely dual adversarial network (DANet),
for real-world image denoising. Different from the traditional MAP framework relied on subjective
pre-assumptions on the noise and image priors, our proposed method
focuses on learning the joint distribution directly from data. To estimate the joint distribution,
we attempt to approximate it by its two different factorized forms using an dual adversarial manner,
which correspondes to two tasks, i.e., noise removal and noise generation.
For assessing the quality of synthetic noisy image, we have designed two applicable metrics, to
the best of our knowledge, for the first time.
The proposed DANet intrinsically provides a general methodology to facilitate the study of other low-level
vision tasks, such as super-resolution and deblurring.
Comprehensive experiments have demonstrated the superiority
of DANet as compared with state-of-the-art methods specifically designed for both the noise removal and
noise generation tasks.

\vspace{2mm}\noindent\textbf{Acknowledgements:} This research was supported by National Key R\&D Program of
China (2018YFB1004300) and the China NSFC project under contracts 11690011, 61721002, U1811461, and Hong Kong
RGC RIF grant (R5001-18).



\clearpage
%
%
\bibliographystyle{splncs04}
\bibliography{egbib-danet}
\end{document}